\documentclass[10pt,twocolumn,letterpaper]{article}

\usepackage[pagenumbers]{cvpr} 

\usepackage{graphicx}
\usepackage{amsmath}
\usepackage{amssymb}
\usepackage{booktabs}
\usepackage[dvipsnames]{xcolor}
\usepackage{xcolor,colortbl}
\usepackage{color, colortbl}
\usepackage{xcolor}
\newcommand{\tableCellHeight}{1}
\newcommand{\tabstyle}[1]{
  \setlength{\tabcolsep}{#1}
  \renewcommand{\arraystretch}{\tableCellHeight}
  \centering
  \small
}
\usepackage{sidecap}
\newcommand{\tablestyle}[2]{\setlength{\tabcolsep}{#1}\renewcommand{\arraystretch}{#2}\centering\footnotesize}
\usepackage{float}
\definecolor{purple}{RGB}{230, 227, 254}
\definecolor{lightgreen}{RGB}{238, 252, 241}
\definecolor{lightred}{RGB}{231, 187, 187}
\definecolor{darkred}{RGB}{198, 129, 129}

\definecolor{tabhighlight}{HTML}{e5e5e5}

\newcommand{\rotbox}[1]{\rotatebox{55}{#1}}
\usepackage{graphicx, amsmath, amssymb, caption, subcaption, multirow, overpic}
\definecolor{tabhighlight}{HTML}{e5e5e5}
\definecolor{citecolor}{HTML}{0071bc}

\definecolor{cvprblue}{rgb}{0.21,0.49,0.74}
\usepackage[pagebackref,breaklinks,colorlinks,allcolors=cvprblue]{hyperref}

\def\paperID{*****} 
\def\confName{CVPR}
\def\confYear{2025}

\title{MSGCoOp: Multiple Semantic-Guided Context Optimization for Few-Shot Learning}

\author{Zhaolong Wang, Tongfeng Sun, Mingzheng Du, Yachao Huang\\
School of Computer Science and Technology, China University of Mining and Technology\\
{\tt\small \{zhaolongwang, suntf, mingzhengdu, yachaohuang\}@cumt.edu.cn}
}

\begin{document}
\maketitle
\begin{abstract}
Vision-language pre-trained models (VLMs) such as CLIP have demonstrated remarkable zero-shot generalization, and prompt learning has emerged as an efficient alternative to full fine-tuning. However, existing methods often struggle with generalization to novel classes, a phenomenon attributed to overfitting on seen classes and forgetting general knowledge. Furthermore, recent approaches that improve generalization often introduce complex architectures or heavy computational overhead. In this paper, we propose a Multiple Semantic-Guided Context Optimization (MSGCoOp) framework to enhance few-shot generalization while maintaining computational efficiency. Our approach leverages an ensemble of parallel learnable context vectors to capture diverse semantic aspects. To enrich these prompts, we introduce a semantic guidance mechanism that aligns them with comprehensive class descriptions automatically generated by a Large Language Model (LLM). Furthermore, a diversity regularization loss encourages the prompts to learn complementary and orthogonal features, preventing them from collapsing into redundant representations. Extensive experiments on 11 benchmark datasets show that MSGCoOp significantly improves performance on base-to-novel generalization, achieving an average harmonic mean improvement of 1.10\% over the strong KgCoOp baseline. Our method also demonstrates enhanced robustness in cross-domain generalization tasks. Our code is avaliable at: \href{https://github.com/Rain-Bus/MSGCoOp}{https://github.com/Rain-Bus/MSGCoOp}.
\end{abstract}    
\section{Introduction}
\label{sec:intro}

Vision-language pre-trained models (VLMs), such as CLIP\cite{radford2021learning} and ALIGN\cite{jia2021scaling}, have revolutionized few-shot and zero-shot image classification by learning transferable representations from web-scale datasets of image-text pairs. These models employ a dual-encoder architecture, comprising an image encoder and a text encoder, trained through contrastive learning to align visual and textual embeddings in a shared feature space. This enables robust generalization to diverse downstream tasks, where image classification is often performed by comparing image embeddings against class-specific text embeddings derived from predefined textual templates (e.g., ``a photo of a \{class name\}''). Such an approach leverages the inherent few-shot capabilities of VLMs, allowing them to recognize novel visual concepts using minimal labeled data.

However, directly adapting these pre-trained models via fine-tuning presents significant challenges. Full-model fine-tuning\cite{gao2024clip, song2023meta} is computationally intensive and risks catastrophic forgetting, degrading the valuable knowledge acquired during large-scale pre-training while potentially overfitting to small downstream datasets. As a result, prompt-based methods(Table~\ref{tab:prompting_methods}) have become popular for improving text representations without expensive fine-tuning. Prompts are textual or continuous inputs appended to the input to guide the model's inference, categorizing into two main types. Hard prompts, or manual prompts, consist of handcrafted word sequences designed to provide human-readable context. Soft prompts are continuous token vectors in a high-dimensional latent space, which can be optimized through training to better adapt to the specific requirements of a task. Compared to hand-crafted textual prompts, soft prompts reduce manual effort and may achieve better few-shot performance. However, their optimization can introduce extra computational cost, and the resulting continuous representations are often more difficult to interpret.

The distinction between soft and hard prompts highlights a core trade-off in prompt engineering for Vision-Language Models (VLMs): balancing the integration of domain-specific knowledge with goals such as computational efficiency, interpretability, and ease of use. As few-shot classification methods increasingly use external knowledge, like linguistic descriptions or structured representations, to enhance text prompts, designing prompts that combine strong performance with practical deployment remains a key challenge for future research.

\begin{figure}[ht]
    \centering
    \includegraphics[width=0.5\textwidth]{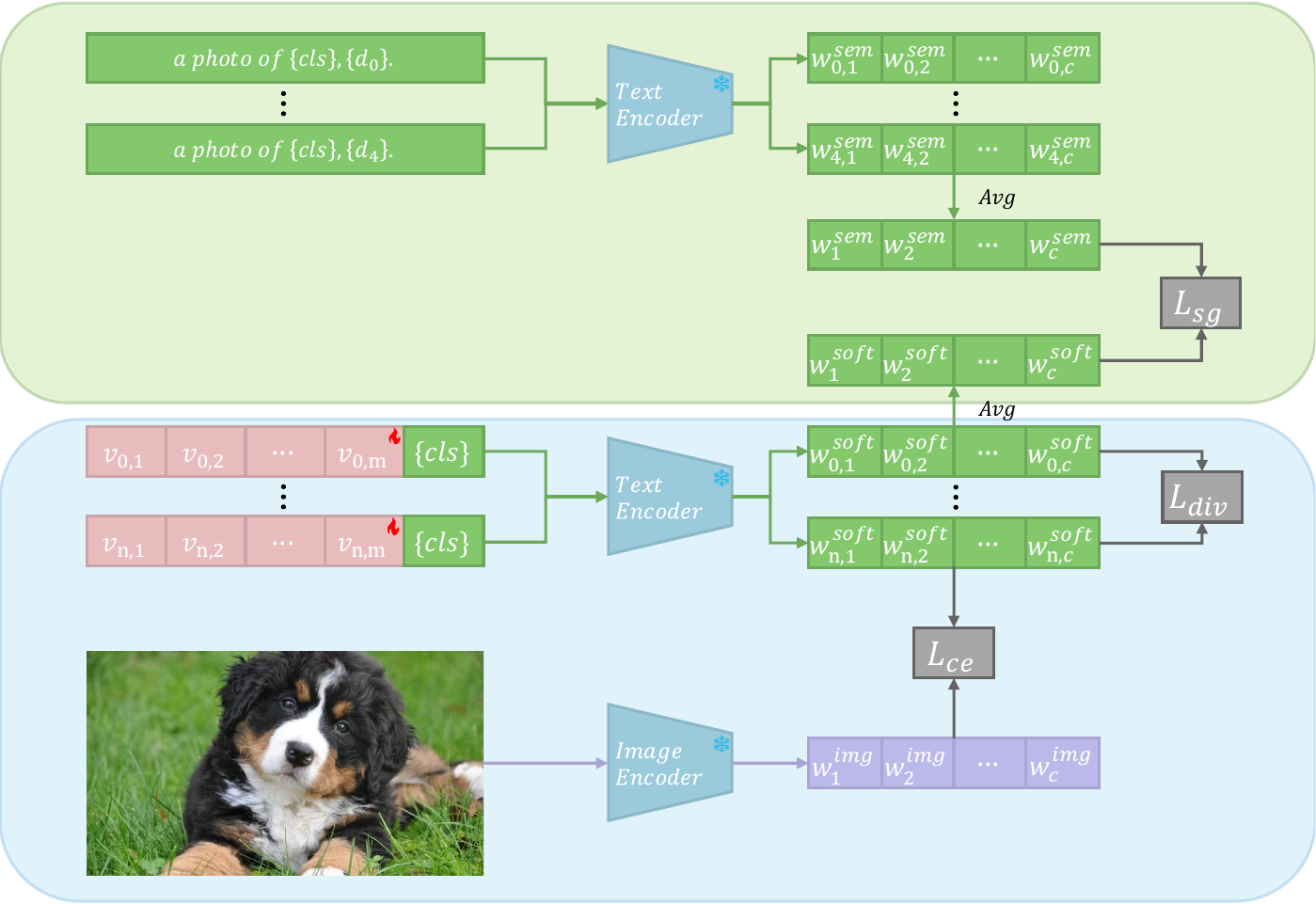}
    \caption{Overview of the MSGCoOp framework. Multi-prompt learning with $N$ parallel context vectors enhances text representations. Semantic guidance and diversity regularization promote alignment with LLM-generated class descriptions and encourage orthogonal prompt features. The frozen CLIP encoders extract image and text embeddings, combined at the logit level for classification.}
    \label{fig:fram}
\end{figure}

\paragraph{Our approach} We propose MSGCoOp, a method for prompt design in CLIP-based few-shot learning. MSGCoOp uses $N$ parallel context vectors as prompts, where $N \geq 1$. These parallel prompts increase diversity while keeping the architecture simple and efficient. MSGCoOp does not add extra trainable network layers and avoids complex coupling mechanisms. Prompt features are fused at the text feature level to enhance classification. To improve text representations, we use class descriptions generated by large language models. This reduces the need for manual prompt engineering and brings in richer semantic information. To encourage diversity among prompts, we add a regularization loss that pushes different context vectors to be orthogonal. This helps each prompt capture different class characteristics.

MSGCoOp shows improved performance on both base-to-novel and cross-domain generalization tasks. On base-to-novel generalization, MSGCoOp increases the average HM accuracy by 1.10\% over the KgCoOp baseline (Table~\ref{table:comparision_with_kgcoop}). For cross-domain generalization, MSGCoOp achieves a 0.30\% gain compared to KgCoOp, and 0.14\% over the multi-modal prompt method MaPLe (Table~\ref{tab:dg}). These results suggest that integrating multi-prompt learning and class descriptions generated by large language models provides richer semantic information for prompt learning. Ablation studies show that semantic-guided prompting improves accuracy by 0.82\% (Table~\ref{tab:abl_LLM}). The multi-prompt approach further brings a 0.58\% increase over single-prompt baselines (Fig.~\ref{fig:abl_N}).

\paragraph{Contributions}
\begin{itemize}
    \item A textual multi-prompt learning framework with parallel context vectors and ensemble fusion.
    \item Automatic semantic enrichment using class descriptions from large language models.
    \item A diversity regularization loss to promote orthogonal and complementary prompts.
    \item Extensive experiments showing improved performance on base-to-novel and cross-domain generalization tasks.
\end{itemize}
\section{Related Works}
\label{sec:relw}

Our work builds upon recent advances in large language models, vision-language pretraining, and prompt learning techniques. We review the most relevant literature across these domains.

\subsection{Large Language Models}

Large Language Models (LLMs) like GPT-3~\cite{brown2020language}, GPT-4~\cite{achiam2023gpt}, and Deepseek-V3~\cite{liu2024deepseek} have transformed natural language processing and are increasingly vital in vision-language tasks. Pre-trained on vast corpora, these models excel in few-shot learning and semantic generation, leveraging transformer-based architectures with billions of parameters to capture intricate linguistic patterns and world knowledge.

In computer vision, LLMs enhance semantic representations in multimodal systems. For instance, Flamingo~\cite{alayrac2022flamingo} uses frozen LLMs with cross-attention to enable few-shot visual question answering, while BLIP-2~\cite{li2023blip} connects vision encoders and LLMs via a Q-Former module for superior image captioning and reasoning.LLaVA~\cite{liu2023llava} aligns a visual encoder with a pretrained LLM through a lightweight projection layer and instruction tuning, enabling flexible image-based question answering and conversation.

\subsection{Vision-Language Models}

Vision-Language Models (VLMs) have emerged as powerful tools for bridging visual and textual modalities through joint representation learning. CLIP~\cite{radford2021learning} pioneered this approach by training dual encoders on 400M image-text pairs using contrastive learning, demonstrating remarkable zero-shot generalization abilities. Following CLIP's success, several works have advanced the field through various improvements, including ALIGN~\cite{jia2021scaling}, Florence~\cite{yuan2021florence}, FILIP~\cite{yao2021filip}, and SLIP~\cite{mu2022slip}.

Recent developments have further enhanced VLMs through novel architectures and training objectives, such as UniCL~\cite{yang2022unified}, KLITE~\cite{shen2022k}, and REACT~\cite{liu2023learning}. These approaches demonstrate the versatility of VLMs across different vision-language tasks while maintaining strong generalization capabilities.

\subsection{Prompt Learning for VLMs}

\begin{table*}
    \centering
    \renewcommand{\arraystretch}{1.2}
    \begin{tabular}{lccp{7.2cm}}
        \toprule
        Method                  & Prompting & Contribution \\
        \midrule
        CLIP~\cite{radford2021learning}       & No        & Zero-shot baseline. \\
        CoOp~\cite{zhou2022learning}          & T         & First prompt learning method for CLIP. \\
        CoCoOp~\cite{zhou2022conditional}     & T + (V)   & Image-conditional textual prompts for better adaptation. \\
        AAPL~\cite{kim2024aapl}               & T + (V)   & Attribute-aware prompts disentangled from augmentation noise. \\
        MaPLe~\cite{khattak2023maple}         & T + V     & Jointly applies textual and visual prompts. \\
        ProGrad~\cite{zhu2023prompt}          & T         & Applying KL loss for multiple logits. \\
        KgCoOp~\cite{yao2023visual}           & T         & Knowledge-guided prompt optimization. \\
        MSGCoOp (Ours)                        & T         & Multi-prompt semantic-guide prompt optimization. \\
        \bottomrule
    \end{tabular}
    \caption{
        Summary of representative prompt learning methods for VLMs. 
        \textbf{T}: textual prompt. 
        \textbf{T + (V)}: text prompt merged with visual information (via conditioning or fusion).
        \textbf{T + V}: textual and visual prompts used in parallel.
    }
    \label{tab:prompting_methods}
\end{table*}

Prompt learning has emerged as an effective approach for adapting vision-language models to downstream tasks without extensive fine-tuning. linear probing often lead to degraded performance or limited zero-shot capability, prompt learning offers a more balanced solution by optimizing continuous prompt vectors in an end-to-end manner.

Context Optimization (CoOp)~\cite{zhou2022learning} pioneered learnable soft prompts for CLIP, replacing hand-crafted templates with trainable context vectors to achieve better few-shot transfer. However, CoOp's fixed prompt strategy showed limitations in generalizing to novel classes. To address this, Conditional Context Optimization (CoCoOp)~\cite{zhou2022conditional} introduced image-conditional prompts by explicitly conditioning prompt generation on image features, improving adaptation to unseen categories at the cost of increased computational overhead.

Recent works have explored various approaches to enhance prompt learning effectiveness. Prompt-aligned Gradient (ProGrad)~\cite{zhu2023prompt} proposed a novel optimization strategy that only updates prompts in directions aligned with general knowledge from zero-shot CLIP predictions, preventing catastrophic forgetting during fine-tuning. Knowledge-guided Context Optimization (KgCoOp)~\cite{yao2023visual} addressed the forgetting issue from a different angle by minimizing the discrepancy between learnable prompts and hand-crafted prompts, achieving better generalization with improved training efficiency.

MaPLe~\cite{khattak2023maple} introduces joint optimization of both textual and visual prompts, inserting learnable tokens into the text and image encoders simultaneously. This approach enables richer cross-modal interaction and improves adaptation to downstream tasks under few-shot settings, outperforming prior methods with stronger generalization to unseen classes.

Adding Attributes to Prompt Learning (AAPL)~\cite{kim2024aapl} introduced adversarial token embedding to disentangle low-level visual augmentation features from high-level class information. By decomposing attribute-specific features through delta meta tokens and employing adversarial triplet loss, AAPL guides the learnable context to effectively focus on high-level features for unseen classes. 

Our method is inspired by KgCoOp. Previous works often introduce heavy computational burdens. For example, CoCoOp relies on image-conditional prompts, AAPL uses adversarial training, and MaPLe requires joint optimization of both encoders. These designs increase the complexity of the framework. In contrast, our approach adopts a simple multi-prompt framework guided by semantic information. We integrate ensemble learning and leverage semantic cues from large language models. This design improves performance while keeping computational costs close to CoOp.

\subsection{Prompt Ensemble}
\label{sec:prompt_ensemble}

Ensemble methods in prompt learning can improve model robustness and performance. Unlike traditional model ensembling that requires multiple model instances, prompt ensembles offer a lightweight alternative by maintaining multiple prompts within a single model. Early work by \cite{schick2020exploiting} demonstrated that averaging predictions from different prompt variants could enhance performance in language tasks, inspiring similar approaches in vision-language domains.

Building on this foundation, the concept of ensembling has been naturally extended to prompt learning for VLMs, where combining predictions from multiple diverse prompts can mitigate individual prompt biases and improve generalization. The simplest form, employed in the original CLIP paper~\cite{radford2021learning}, involves averaging predictions over hand-crafted prompts (e.g., ``a photo of a \{class\}'', ``an image of a \{class\}''). While this manual approach may not discover optimal prompt variations.

Our method proposes an efficient approach for prompt ensembling. We learn $N$ distinct and parallel context vectors with a diversity regularization loss that encourages orthogonality in the feature space. This helps each prompt capture different semantic information, preventing them from collapsing to similar representations. We further combine this diversity mechanism with efficient feature fusion and semantic guidance from large language models. MSGCoOp builds an effective ensemble that benefits from external knowledge, while remaining simpler and more efficient.

\section{Method}
\label{sec:method}

We propose a novel prompt learning framework designed to enhance few-shot generalization in CLIP-based models. Our approach introduces \textbf{multi-prompt learning}, \textbf{semantic-rich class description guide}, and a \textbf{diversity regularization loss} to boost both expressiveness and robustness of learned prompts. Figure~\ref{fig:fram} illustrates the overall pipeline of our proposed method. During training, only the prompt-related components are optimized, while the CLIP encoders remain frozen.

\subsection{Overview of CLIP-based Prompt Learning}

We leverage the Contrastive Language-Image Pretraining (CLIP) framework, which consists of a vision encoder $\phi(\cdot)$ and a text encoder $\theta(\cdot)$ trained to align image-text pairs in a shared embedding space. For a $N_c$ classes classification task, we denote the class names as $\{\mathrm{cls}_1, ..., \mathrm{cls}_i, ..., \mathrm{cls}_{N_c}\}$, CLIP's zero-shot capability is realized by class-specific textual prompts (e.g. ``a photo of a [CLASS]''). The textual embeddings are computed as $\mathbf{w}_i^{\mathrm{txt}} = \theta(\text{``a photo of a $\mathrm{cls}_i,$''})$ for each class. Given an input image, its visual feature representation is extracted as $\mathbf{w}^{\mathrm{img}} = \phi(I)$, where $I$ denotes the input image. The prediction scores are derived from the cosine similarity between $\mathbf{w}^{\mathrm{img}}$ and each textual embedding $\mathbf{w}_i^{\mathrm{txt}}$.

To improve performance in low-data regimes, prompt tuning methods (e.g. CoOp) replace fixed text templates with learnable continuous vectors. However, such approaches may suffer from overfitting or limited diversity due to the use of a single prompt representation per class.

\subsection{Semantic-rich Description Generation}
To enrich the semantic representation of each class, we use LLM to generate comprehensive class descriptions that capture distinctive visual characteristics. Unlike previous approaches that rely on simple templates or manual annotations, our method automatically generates multiple diverse descriptions for each class through structured prompting.

\begin{figure}[ht]
    \centering
    \includegraphics[width=0.5\textwidth]{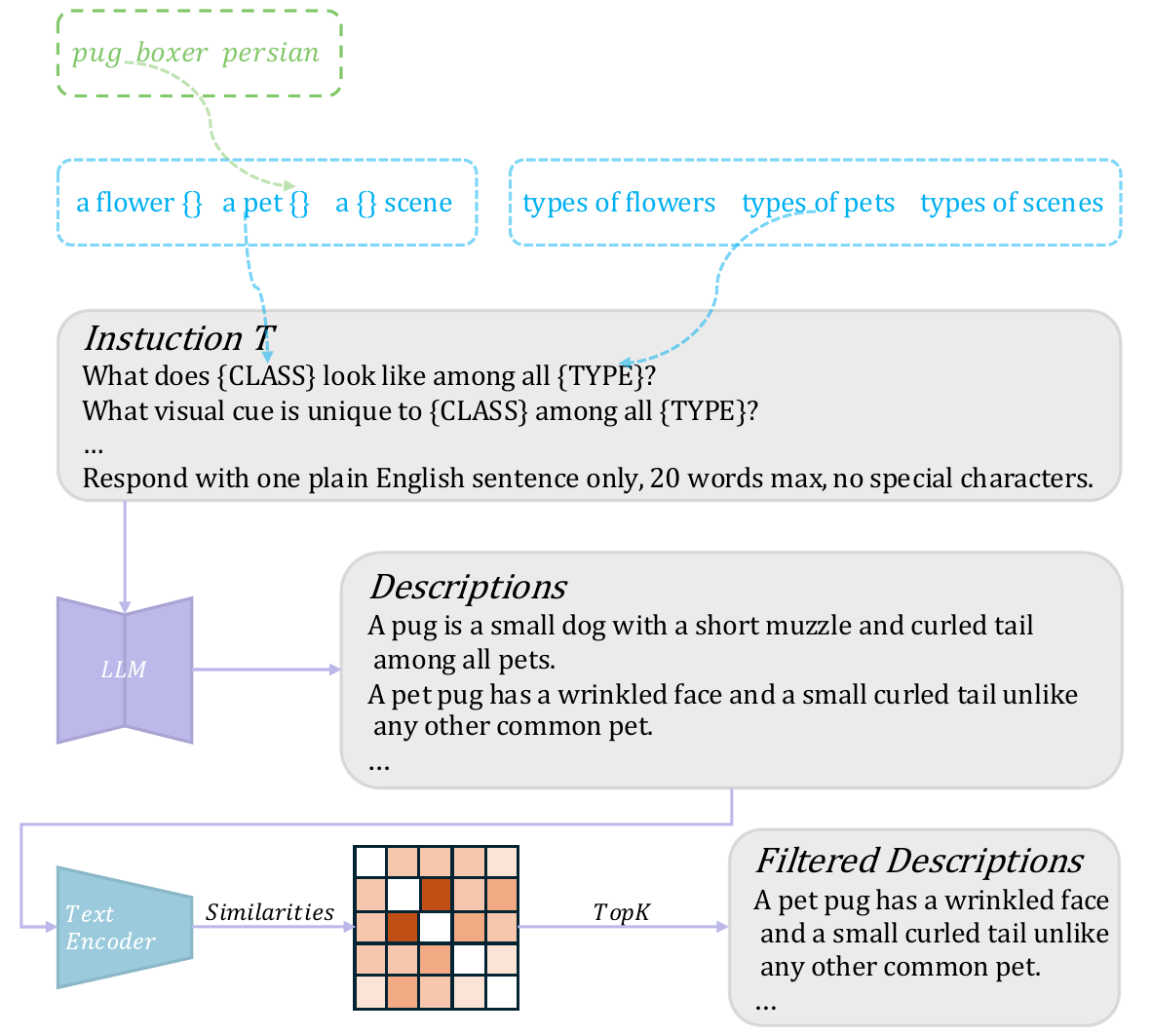}
    \caption{Pipeline for generating semantic-rich class descriptions using a large language model (LLM). Structured prompts guide the LLM to produce multiple concise descriptions focusing on visual discriminative features. A text encoder filters and selects the top-$k$ most consistent descriptions based on similarity scores, providing semantically enriched inputs for multi-prompt learning.}
    \label{fig:desc_gen}
\end{figure}

\paragraph{LLM Prompt Design} We design a set of complementary prompt templates that focus on visual attributes and discriminative features:
\begin{itemize}
    \item What does [CLASS] look like among all [CATEGORY]?
    \item What visual cue is unique to [CLASS] among all [CATEGORY]?
    \item What are the distinct features of [CLASS] for recognition among all [CATEGORY]?
    \item How can you identify [CLASS] in appearance among all [CATEGORY]?
    \item What are the differences between [CLASS] and other [CATEGORY] in appearance?
\end{itemize}
where [CLASS] represents the target class name and [CATEGORY] indicates the semantic super-category or domain of the dataset(e.g. ``types of pets" for OxfordPets). This design constrains the LLM to generate descriptions that emphasize intra-domain discriminative features.

\paragraph{Description Generation Pipeline} For each class, we first instantiate the prompt templates with corresponding [CLASS] and [CATEGORY]. These prompts are then passed to the LLM with a system prompt that positions it as an expert in visual feature analysis. To maintain consistency and conciseness, we constrain the LLM to generate single-sentence descriptions with a maximum of 20 words per response.

The generated descriptions $D_{cls} = \{d_1, ..., d_K\}$ for class $cls$ are then filtered to select the most semantically relevant subset, where $K$ is the total number of raw descriptions generated by the LLM for each class. We compute the relevance scores using CLIP's text encoder:
\begin{equation}
    s_{i,j} = \mathrm{sim}(\theta(d_i), \theta(d_j))
\end{equation}
where $\mathrm{sim}(\cdot,\cdot)$ denotes cosine similarity between two vectors. For each description $d_i$, we calculate its average similarity with other descriptions:
\begin{equation}
    \bar{s}_i = \frac{1}{K-1} \sum_{j \ne i} s_{i,j}
\end{equation}

The top-$k$ descriptions with highest average similarity scores are selected to form the final description set $\hat{D}_{cls}$ for each class, which retain the most representative and consistent semantic concepts for each class, reducing noise from potentially irrelevant or outlier descriptions generated by the LLM. These descriptions are then used to construct class-specific prompt embeddings that guide the multi-prompt learning process.

\subsection{Multi-Prompt Learning Mechanism}

To address the limitations of single-prompt representations in few-shot image classification, we introduce a multi-prompt learning mechanism that enables the model to efficiently capture diverse class semantics through an ensemble of learnable prompt vectors. Our approach extends the traditional prompt learning framework by optimizing $N$ parallel context prompts per class, thereby promoting diversity and robustness in downstream classification tasks.

\paragraph{Prompt Construction.}
We utilize a textual template to generate the initial prompt for each class, e.g. ``a photo of \(\mathrm{cls}_i\).'' In standard prompt tuning methods such as CoOp~\cite{zhou2022learning}, a single learnable context vector is concatenated with the class name to form the input to the CLIP text encoder. Our multi-prompt framework constructs $N$ independent context vectors
\begin{equation}
    \mathbf{p}_{i,n} = \big[\mathbf{v}_n, \mathbf{c}_i\big]
\end{equation}
where $\mathbf{v}_n = \big[v_{n,1}, v_{n,2}, \dots, v_{n,m} \big] \in \mathbb{R}^{M \times d}$ is the $n$-th learnable context vector of length $M$ and embedding dimension $d$, and $\mathbf{c}_i$ is the token embedding of the class name $\mathrm{cls}_i$. Each prompt $\mathbf{p}_{i,n}$ is then tokenized and encoded via the frozen CLIP text encoder $\theta(\cdot)$ to obtain a set of $N$ prompt-derived text features for class $\mathrm{cls}_i$:
\begin{equation}
    \mathbf{w}_{i,n}^{\mathrm{soft}} = \theta(\mathbf{p}_{i,n}), \quad n = 1, ..., N
\end{equation}

\paragraph{Ensemble Fusion.}
At inference, given an image $I$ and its feature vector $\mathbf{w}^{\mathrm{img}} = \phi(I)$ from the frozen CLIP image encoder, the model computes the cosine similarity between the image feature and each prompt-derived text feature for each class. The score of $\mathrm{cls}_i$ is obtained by averaging the logits across all $N$ prompts:
\begin{equation}
    s_i = \frac{1}{N} \sum_{n=1}^{N} \mathrm{sim} \big( \mathbf{w}^{\mathrm{img}}, \mathbf{w}_{i,n}^{\mathrm{soft}} \big)
\end{equation}
The final predicted label is $\hat{y} = \arg \max_{i} s_i$.

\paragraph{Objective.}
During training, only the context vectors $\{\mathbf{v}_{n}\}_{n=1}^{N}$ are optimized via gradient descent, while the image and text encoders remain frozen. The cross-entropy loss is computed over the average class probabilities derived from the ensemble logits:
\begin{equation}
   \mathcal{L}_{\mathrm{ce}} = -\frac{1}{B} \sum_{j=1}^{B} \sum_{i=1}^{N_c} y_{j,i} \log \left( \frac{\exp(s_{j,i})}{\sum_{k=1}^{N_c} \exp(s_{j,k})} \right)
\end{equation}
where $B$ is the batch size, $y_{j,i}$ is the one-hot ground truth label for sample $j$ and class $i$, and $s_{j,i}$ is the ensemble logits for sample $j$ and class $i$ as defined in Eq.5. Unlike multi-modal or conditional prompt learning approaches~\cite{zhou2022conditional, khattak2023maple}, our method maintains architectural simplicity and computational efficiency, as all prompt variants are encoded in parallel and combined at the logit level.

\subsection{Semantic Guide Regularization}
To leverage the rich semantic information from LLM-generated descriptions, we introduce a semantic guidance mechanism that encourages learned prompts to maintain consistency with these descriptions while allowing for task-specific adaptation. For each class $cls$, we first encode its descriptions $\hat{D}_{cls}$ using the CLIP text encoder to obtain semantic reference embeddings:

\begin{equation}
    \mathbf{w}^{\mathrm{sem}} = \frac{1}{|\hat{D}_{cls}|} \sum_{d \in \hat{D}_{cls}} \theta(d)
\end{equation}

We then define a semantic guidance loss that minimizes the divergence between the averaged prompt embeddings and semantic reference embeddings:

\begin{equation}
    \mathcal{L}_{\mathrm{sg}} = \frac{1}{N_c} \sum_{i=1}^{N_c} \left[ 1 - \mathrm{sim}\big( \bar{\mathbf{w}}_i^{\mathrm{soft}}, \mathbf{w}_i^{\mathrm{sem}} \big) \right]
\end{equation}

where $\bar{\mathbf{w}}_i^{\mathrm{soft}} = \frac{1}{N} \sum_{n=1}^{N} \mathbf{w}_{i,n}^{\mathrm{soft}}$ is the mean of learned prompt embeddings for class $\mathrm{cls}_i$. This regularization term ensures that the ensemble of prompts maintains semantic alignment with the class-specific visual characteristics captured in the LLM-generated descriptions.

\subsection{Diversity Regularization}
To prevent the multiple prompts from converging to similar representations and to encourage the capture of complementary class characteristics, we introduce a diversity regularization loss. For each class $\mathrm{cls}_i$, we compute the pairwise similarities between its $N$ prompt embeddings and penalize high similarity scores:

\begin{equation}
    \mathcal{L}_{\mathrm{div}} = \frac{1}{N_c} \sum_{i=1}^{N_c} \frac{1}{N(N-1)} \sum_{m=1}^{N} \sum_{n \neq m} \mathrm{cos}^2 \big(\mathbf{w}_{i,m}^{\mathrm{soft}}, \mathbf{w}_{i,n}^{\mathrm{soft}} \big)
\end{equation}

This orthogonality-promoting regularization encourages each prompt to focus on different discriminative aspects of the class, leading to more robust ensemble predictions. We choose squared cosine similarity as a lightweight and bounded surrogate to promote orthogonality, which empirically leads to stable optimization.

\subsection{Final Objective}
The full training objective combines classification, diversity, and alignment losses:

\begin{equation}
    \mathcal{L}_{\mathrm{total}} = \mathcal{L}_{\mathrm{ce}} + \lambda_{\mathrm{sg}} \mathcal{L}_{\mathrm{sg}} + \lambda_{\mathrm{div}} \mathcal{L}_{\mathrm{div}}
\end{equation}

where $\mathcal{L}_{\text{ce}}$ is the standard cross-entropy loss computed over ensemble predictions, and $\lambda_{\text{sg}}$, $\lambda_{\text{div}}$ are weighting factors balancing the regularization terms.

This multi-objective formulation enables our model to effectively learn expressive, diverse, and semantically aligned prompts for robust few-shot classification.

\section{Experiments}
\label{sec:exp}

\subsection{Experimental Settings}
\label{sec:exp_set}
\noindent\textbf{Base-to-Novel Generalization:} 
We evaluate the generalization capability of MSGCoOp following the standard few-shot setting. Each dataset is split into base classes (used for training) and novel classes (held-out for testing). The model is trained on $K$-shot examples per base class and evaluated on both base and novel classes under a zero-shot transfer protocol. We report base accuracy (B), novel accuracy (N), and their harmonic mean (HM).

\noindent\textbf{Cross-Dataset Generalization:} 
To assess the cross-dataset generalization, we train our model on all 1000 classes of ImageNet in a 16-shot setting and directly evaluate it on 10 downstream datasets without any dataset-specific fine-tuning. This measures the model's ability to generalize to diverse recognition tasks beyond the source domain.

\noindent\textbf{Domain Generalization:} 
We evaluate robustness to domain shifts by treating ImageNet as the source domain and testing on four out-of-distribution variants: ImageNetV2~\cite{recht2019imagenet} (natural distribution shift), ImageNet-Sketch~\cite{wang2019learning} (sketch images), ImageNet-A~\cite{hendrycks2021natural} (adversarial examples), and ImageNet-R~\cite{hendrycks2021many} (artistic renditions). Models are trained on ImageNet and directly evaluated on target domains.

\noindent\textbf{Datasets:} 
Following CoOp~\cite{zhou2022learning}, and KgCoOp~\cite{yao2023visual}, we use 11 image classification datasets covering diverse recognition tasks:
\begin{itemize}[itemsep=1pt,topsep=0pt,parsep=0pt]
    \item \textit{Generic object classification:} ImageNet~\cite{deng2009imagenet}, Caltech101~\cite{fei2004learning}
    \item \textit{Fine-grained visual categorization:} OxfordPets~\cite{parkhi2012cats}, StanfordCars~\cite{krause20133d}, Flowers102~\cite{nilsback2008automated}, Food101~\cite{bossard2014food}, FGVCAircraft~\cite{maji2013fine}
    \item \textit{Specialized recognition:} EuroSAT~\cite{helber2019eurosat} (satellite), UCF101~\cite{soomro2012ucf101} (action), DTD~\cite{cimpoi2014describing} (texture), SUN397~\cite{xiao2010sun} (scenes)
\end{itemize}
For domain generalization, we use ImageNet as source and its four variants as target domains.

\noindent\textbf{Implementation Details:} 
We use the CLIP ViT-B/16 backbone with $d_l=512$ text embedding dimension and $d_v=768$ vision embedding dimension. All models are trained in 16-shot setting with batch size 128 for 100 epochs using SGD optimizer with learning rate 0.002, which is same as KgCoOp\cite{yao2023visual}. 

We initialize the context vectors using the template ``a photo of a'' and set the prompt length to 4 tokens. Class descriptions are generated by GPT-4~\cite{achiam2023gpt}, retaining $k=4$ descriptions per class after similarity filtering. We set the hyperparameters $\lambda_{\mathrm{sg}}=8.0$ and $\lambda_{\mathrm{div}}=1.0$ in the loss function. All results are averaged over 3 random seeds. Experiments are conducted on 40-series NVIDIA GPUs provided by \href{https://www.autodl.com/docs/gpu_perf/}{AutoDL}. For the base-to-novel evaluation, we use $N=4$ parallel prompts per class on vGPU-32G instances. For the cross-domain evaluation, we use $N=3$ parallel prompts per class on vGPU-48G instances due to resource constraints.

\subsection{Base-to-Novel Generalization}

\begin{table*}[t!]
\tablestyle{6pt}{0}
\addtolength{\tabcolsep}{-6pt}
    \tabstyle{1.5pt}
    \setlength{\tabcolsep}{6pt}
    \begin{subtable}[t]{.32\textwidth}
    \centering
    \caption{\textbf{Average over 11 datasets}}
    \begin{tabular}{l cc|c}
    \toprule
    & Base & Novel & HM \\
    \midrule
    CLIP & 69.34 & 74.22 & 71.70 \\
    CoOp & \textbf{82.69} & 63.22 & 71.66 \\
    Co-CoOp & 80.47 & 71.69 & 75.83 \\
    AAPL & 80.27 & 72.17 & 76.01 \\
    KgCoOp & 80.73 & 73.36 & 77.00 \\
    \midrule
    \rowcolor{tabhighlight}
    MSGCoOp & 81.40 & \textbf{75.05} & \textbf{78.10} \\
     & \textcolor{MidnightBlue}{{+0.67}} & \textcolor{MidnightBlue}{{+1.69}} & \textcolor{MidnightBlue}{{+1.10}} \\
    \bottomrule
    \end{tabular}
    \end{subtable}
    \vspace{1em}
    \begin{subtable}[t]{.32\textwidth}
    \centering
    \caption{ImageNet.}
    \begin{tabular}{l cc|c}
    \toprule
    & Base & Novel & HM \\
    \midrule
    CLIP & 72.43 & 68.14 & 70.22 \\
    CoOp & 76.47 & 67.88 & 71.92\\
    Co-CoOp & 75.98 & 70.43 & 73.10 \\
    AAPL & \textbf{76.53} & \textbf{70.57} & \textbf{73.43} \\
    KgCoOp & 75.83 & 69.96 & 72.78 \\
    \midrule
    \rowcolor{tabhighlight}
    MSGCoOp & 76.46 & 70.45 & 73.31 \\
       & \textcolor{MidnightBlue}{{+0.63}} & \textcolor{MidnightBlue}{{+0.49}} & \textcolor{MidnightBlue}{{+0.53}} \\
    \bottomrule
    \end{tabular}
    \end{subtable}
    ~
    \begin{subtable}[t]{.32\textwidth}
    \centering
    \caption{Caltech101}
    \begin{tabular}{l cc|c}
    \toprule
    & Base & Novel & HM \\
    \midrule
    CLIP & 96.84 & 94.00 & 95.40 \\
    CoOp & 98.00 & 89.81 & 93.73 \\
    Co-CoOp & 97.96 & 93.81 & 95.84 \\
    AAPL & 97.87 & 95.10 & 96.46 \\
    KgCoOp & 97.72 & 94.39 & 96.03 \\
    \midrule
    \rowcolor{tabhighlight}
    MSGCoOp & \textbf{98.08} & \textbf{95.81} & \textbf{96.91} \\
      & \textcolor{MidnightBlue}{{+0.36}} & \textcolor{MidnightBlue}{{+1.42}} & \textcolor{MidnightBlue}{{+0.88}} \\
    \bottomrule
    \end{tabular}
    \end{subtable}
    ~
    \begin{subtable}[t]{.32\textwidth}
    \centering
    \caption{OxfordPets}
    \begin{tabular}{l cc|c}
    \toprule
    & Base & Novel & HM \\
    \midrule
    CLIP & 91.17 & 97.26 & 94.12 \\
    CoOp & 93.67 & 95.29 & 94.47 \\
    Co-CoOp & 95.20 & 97.69 & 96.43 \\
    AAPL & \textbf{95.63} & 97.40 & 96.51 \\
    KgCoOp & 94.65 & 97.76 & 96.18 \\
    \midrule
        \rowcolor{tabhighlight}
    MSGCoOp & 95.62 & \textbf{97.89} & \textbf{96.70} \\
      & \textcolor{MidnightBlue}{{+0.97}} & \textcolor{MidnightBlue}{{+0.13}} & \textcolor{MidnightBlue}{{+0.52}} \\
    \bottomrule
    \end{tabular}
    \end{subtable}
    \vspace{1em}
    \begin{subtable}[t]{.32\textwidth}
    \centering
    \caption{StanfordCars}
    \begin{tabular}{l cc|c}
    \toprule
    & Base & Novel & HM \\
    \midrule
    CLIP & 63.37 & 74.89 & 68.65 \\
    CoOp & \textbf{78.12} & 60.40 & 68.13 \\
    Co-CoOp & 70.49 & 73.59 & 72.01 \\
    AAPL & 70.33 & 73.50 & 71.88 \\
    KgCoOp & 71.76 & \textbf{75.04} & \textbf{73.36} \\
    \midrule
        \rowcolor{tabhighlight}
    MSGCoOp & 71.22 & 74.56 & 72.85 \\
    & \textcolor{Bittersweet}{-0.54} & \textcolor{Bittersweet}{-0.48} & \textcolor{Bittersweet}{-0.51} \\
    \bottomrule
    \end{tabular}
    \end{subtable}
    ~
    \begin{subtable}[t]{.32\textwidth}
    \centering
    \caption{Flowers102}
    \begin{tabular}{l cc|c}
    \toprule
    & Base & Novel & HM \\
    \midrule
    CLIP & 72.08 & \textbf{77.80} & 74.83 \\
    CoOp & \textbf{97.60} & 59.67 & 74.06 \\
    Co-CoOp & 94.87 & 71.75 & 81.71 \\
    AAPL & 95.10 & 70.63 & 81.06 \\
    KgCoOp & 95.00 & 74.73 & 83.65 \\
    \midrule
        \rowcolor{tabhighlight}
    MSGCoOp & 96.20 & 75.91 & \textbf{84.84} \\
      & \textcolor{MidnightBlue}{{+1.20}} & \textcolor{MidnightBlue}{{+1.18}} & \textcolor{MidnightBlue}{{+1.19}} \\
    \bottomrule
    \end{tabular}
    \end{subtable}
    ~
    \begin{subtable}[t]{.32\textwidth}
    \centering
    \caption{Food101}
    \begin{tabular}{l cc|c}
    \toprule
    & Base & Novel & HM \\
    \midrule
    CLIP & 90.10 & 91.22 & 90.66 \\
    CoOp & 88.33 & 82.26 & 85.19 \\
    Co-CoOp & 90.70 & 91.29 & 90.99 \\
    AAPL & 90.70 & 91.60 & 91.15 \\
    KgCoOp & 90.50 & \textbf{91.70} & 91.09 \\
    \midrule
        \rowcolor{tabhighlight}
    MSGCoOp & \textbf{90.72} & 91.62 & \textbf{91.18} \\
      & \textcolor{MidnightBlue}{{+0.22}} & \textcolor{Bittersweet}{{-0.08}} & \textcolor{MidnightBlue}{{+0.09}} \\
    \bottomrule
    \end{tabular}
    \end{subtable}
    \vspace{1em}
    \begin{subtable}[t]{.32\textwidth}
    \centering
    \caption{FGVCAircraft}
    \begin{tabular}{l cc|c}
    \toprule
    & Base & Novel & HM \\
    \midrule
    CLIP & 27.19 & \textbf{36.29} & 31.09 \\
    CoOp & \textbf{40.44} & 22.30 & 28.75 \\
    Co-CoOp & 33.41 & 23.71 & 27.74 \\
    AAPL & 34.07 & 24.17 & 28.28 \\
    KgCoOp & 36.21 & 33.55 & 34.83 \\
    \midrule
        \rowcolor{tabhighlight}
    MSGCoOp & 36.73 & 34.57 & \textbf{35.62} \\
      & \textcolor{MidnightBlue}{{+0.52}} & \textcolor{MidnightBlue}{{+1.02}} & \textcolor{MidnightBlue}{{+0.79}} \\
    \bottomrule
    \end{tabular}
    \end{subtable}
    ~
    \begin{subtable}[t]{.32\textwidth}
    \centering
    \caption{SUN397}
    \begin{tabular}{l cc|c}
    \toprule
    & Base & Novel & HM \\
    \midrule
    CLIP & 69.36 & 75.35 & 72.23 \\
    CoOp & 80.60 & 65.89 & 72.51 \\
    Co-CoOp & 79.74 & 76.86 & 78.27 \\
    AAPL & 79.65 & \textbf{76.90} & 78.25 \\
    KgCoOp  & 80.29 & 76.53 & 78.36 \\
    \midrule
        \rowcolor{tabhighlight}
    MSGCoOp & \textbf{81.38} & 76.20 & \textbf{78.71} \\
      & \textcolor{MidnightBlue}{{+1.09}} & \textcolor{Bittersweet}{{-0.33}} & \textcolor{MidnightBlue}{{+0.35}} \\
    \bottomrule
    \end{tabular}
    \end{subtable}
    ~
    \begin{subtable}[t]{.32\textwidth}
    \centering
    \caption{DTD}
    \begin{tabular}{l cc|c}
    \toprule
    & Base & Novel & HM \\
    \midrule
    CLIP & 53.24 & \textbf{59.90} & 56.37 \\
    CoOp & \textbf{79.44} & 41.18 & 54.24 \\
    Co-CoOp & 77.01 & 56.00 & 64.85 \\
    AAPL & 73.90 & 53.43 & 62.02 \\
    KgCoOp & 77.55 & 54.99 & 64.35 \\
    \midrule
        \rowcolor{tabhighlight}
    MSGCoOp & 79.32 & 56.64 & \textbf{66.09} \\
      & \textcolor{MidnightBlue}{{+1.77}} & \textcolor{MidnightBlue}{{+1.65}} & \textcolor{MidnightBlue}{{+1.74}} \\
    \bottomrule
    \end{tabular}
    \end{subtable}
    ~
    \begin{subtable}[t]{.32\textwidth}
    \centering
    \caption{EuroSAT}
    \begin{tabular}{l cc|c}
    \toprule
    & Base & Novel & HM \\
    \midrule
    CLIP & 56.48 & 64.05 & 60.03 \\
    CoOp & \textbf{92.19} & 54.74 & 68.69 \\
    Co-CoOp & 87.49 & 60.04 & 71.21 \\
    AAPL & 87.00 & 66.30 & 75.25 \\
    KgCoOp & 85.64 & 64.34 & 73.48 \\
    \midrule
        \rowcolor{tabhighlight}
    MSGCoOp & 86.29 & \textbf{74.97} & \textbf{80.23} \\
      & \textcolor{MidnightBlue}{{+0.65}} & \textcolor{MidnightBlue}{{+10.63}} & \textcolor{MidnightBlue}{{+6.75}} \\
    \bottomrule
    \end{tabular}
    \end{subtable}
    ~
    \begin{subtable}[t]{.32\textwidth}
    \centering
    \caption{UCF101}
    \begin{tabular}{l cc|c} 
    \toprule
    & Base & Novel & HM \\
    \midrule
    CLIP & 70.53 & \textbf{77.50} & 73.85 \\
    CoOp & \textbf{84.69} & 56.05 & 67.46 \\
    Co-CoOp & 82.33 & 73.45 & 77.64 \\
    AAPL & 82.20 & 74.27 & 78.03 \\
    KgCoOp & 82.89 & 76.67 & 79.65 \\
    \midrule
        \rowcolor{tabhighlight}
    MSGCoOp & 83.52 & 77.05 & \textbf{80.16} \\
     & \textcolor{MidnightBlue}{{+0.63}} & \textcolor{MidnightBlue}{{+0.38}} & \textcolor{MidnightBlue}{{+0.51}} \\
    \bottomrule
    \end{tabular}
    \end{subtable}
    \caption{\small\textbf{Comparison with previous textual prompt learning methods on base-to-novel generalization}. MSGCoOp learns multiple semantic-guided prompts and demonstrates strong generalization results over existing methods on 11 recognition datasets. Absolute gains over KgCoOp are indicated in \textcolor{MidnightBlue}{blue}.}
    \label{table:comparision_with_kgcoop}
\end{table*}

We evaluate the generalization ability from base classes (seen during training) to novel classes (unseen during training) on 11 recognition datasets.

\textbf{Main Results:} Table~\ref{table:comparision_with_kgcoop} shows results on 11 datasets. MSGCoOp achieves an average base accuracy of 81.40\%, novel accuracy of 75.05\%, and harmonic mean of 78.10\%. Compared to KgCoOp, the gains are +0.67\% on base accuracy, +1.69\% on novel accuracy, and +1.10\% on the harmonic mean.

\textbf{Novel Class Performance:} On novel classes, MSGCoOp achieves higher accuracy than KgCoOp on 8 out of 11 datasets. The average improvement is +1.69\%. The largest gains are on EuroSAT (+10.63\%) and DTD (+1.65\%).

\textbf{Harmonic Mean Analysis:} MSGCoOp obtains a higher harmonic mean than KgCoOp on 10 out of 11 datasets. The average harmonic mean is 78.10\%, which is +1.10\% higher than KgCoOp. The largest improvements are on EuroSAT (+6.75\%), DTD (+1.74\%), and Flowers102 (+1.19\%).

\subsection{Cross-Domain Generalization}
\label{sec: xdo_gen}

\begin{table*}[]
\centering
\small
\caption{Comparison of prompt learning methods for cross-domain generalization with 16-shot setting.}
\label{tab:dg}
\vspace{-0.8em}
\begin{tabular}{l|c|cccc|c}
\toprule
        & Source   & \multicolumn{5}{c}{Target}                             \\\cline{2-7} 
        & ImageNet & ImageNetV2 & ImageNet-Sketch & ImageNet-A & ImageNet-R & Avg. \\
\midrule
CLIP             & 66.73     & 60.83    & 46.15           & 47.77        & 73.96     & 57.17 \\ 
CoOp             & 71.51     & 64.20    & 47.99           & 49.71        & 75.21     & 59.28 \\
CoCoOp           & 71.02     & 64.07    & 48.75           & 50.63        & 76.18     & 59.90 \\
AAPL             & 71.37     & 64.20    & 48.80           & 50.60        & 76.87     & 60.12 \\
MaPLe            & 70.72     & 64.07    & 49.15           & \textbf{50.90} & \textbf{76.98} & 60.27 \\
KgCoOp           & 71.20     & 64.10    & 48.97           & 50.69        & 76.70     & 60.11 \\
\midrule
\rowcolor{tabhighlight}
MSGCoOp          & \textbf{71.59} & \textbf{64.54} & \textbf{49.24} & \textbf{50.90} & 76.94 & \textbf{60.41} \\
\bottomrule
\end{tabular}
\vspace{-0.8em}
\end{table*}

Domain generalization measures a model’s robustness to out-of-distribution (OOD) shifts. This is important for vision-language models used in real-world applications, where the test data often differs from the training data. Following previous work~\cite{zhou2022learning, zhou2022conditional, yao2023visual, zhu2023prompt}, we train models on ImageNet and evaluate them on four ImageNet variants (see Section~\ref{sec:exp_set}) as target domains. No domain-specific adaptation or extra prompt tuning is used.

\textbf{Main Results:} Table~\ref{tab:dg} shows that MSGCoOp achieves the highest average accuracy of 60.41\% across the four target domains. This result is better than all baseline methods. Compared to KgCoOp, which is the best baseline, MSGCoOp improves average target accuracy by 0.30\% (60.41\% vs. 60.11\%), and also increases source domain accuracy (71.59\% vs. 71.20\%). Compared to MaPLe, which uses both textual and visual prompt tuning and achieves 60.27\%, MSGCoOp gives an improvement of 0.14\%.

\textbf{Target Domain Performance:} MSGCoOp outperforms all baselines on every target domain. Compared to KgCoOp, the gains are 0.44\% on ImageNetV2, 0.27\% on ImageNet-Sketch, 0.21\% on ImageNet-A, and 0.24\% on ImageNet-R. Compared to MaPLe, the improvements are 0.47\% on ImageNetV2, 0.09\% on ImageNet-Sketch, 0.00\% on ImageNet-A, and -0.04\% on ImageNet-R.

\subsection{Cross-Dataset Generalization}

\begin{table*}[]
    \caption{Comparison of textual prompt learning methods for cross-dataset generalization with 16-shot setting. $ep$ represents train epoch.}
    \label{tab:dag}
    \tabstyle{4pt}
    \scalebox{0.85}{
    \begin{tabular}{l c ccccccccccc}
    \toprule
    & \textbf{Source} & \multicolumn{11}{c}{\textbf{Target}} \\ \cmidrule(lr){2-2} \cmidrule(lr){3-13}
    & \rotbox{ImageNet} & \rotbox{Caltech101} & \rotbox{OxfordPets} & \rotbox{StanfordCars} & \rotbox{Flowers102} & \rotbox{Food101} & \rotbox{Aircraft} & \rotbox{SUN397} & \rotbox{DTD} & \rotbox{EuroSAT} & \rotbox{UCF101} & \rotbox{\emph{Average}} \\
    \midrule
    CLIP & 66.72 & 92.94 & 89.07 & 65.29 & 71.30 & 86.11 & \textbf{24.87} & 62.62 & 44.56 & \textbf{47.69} & 66.77 & 65.12 \\
    \midrule
    CoOp & 71.51 & 93.70 & 89.14 & 64.51 & 68.71 & 85.30 & 18.47 & 64.15 & 41.92 & 46.39 & 66.55 & 63.88 \\
    Co-CoOp & 71.02 & 94.43 & 90.14 & 65.32 & \textbf{71.88} & 86.06 & 22.94 & 67.36 & 45.73 & 45.37 & 68.21 & 65.74 \\
    AAPL & 71.37 & 94.17 & \textbf{90.73} & 65.10 & 71.67 & 86.00 & 23.03 & 66.80 & 44.80 & 41.83 & \textbf{69.30} & 65.34 \\
    KgCoOp & 70.66 & 93.92 & 89.83 & 65.41 & 70.01 & \textbf{86.36} & 22.51 & 66.16 & \textbf{46.35} & 46.04 & 68.50 & 65.51 \\
    \midrule
    \rowcolor{tabhighlight} MSGCoOp$_{ep5}$ & 70.52 & \textbf{94.66} & 89.56 & \textbf{65.78} & 71.85 & 86.15 & 23.16 & \textbf{67.66} & 45.41 & 46.16 & 68.80 & \textbf{65.92} \\
    \rowcolor{tabhighlight} MSGCoOp$_{ep100}$ & \textbf{71.59} & 93.29 & 87.90 & 65.10 & 69.87 & 85.60 & 19.14 & 65.14 & 42.59 & 45.59 & 67.51 & 64.17 \\
    \bottomrule
    \end{tabular}}
\end{table*}

We evaluate the generalization ability of MSGCoOp by training on ImageNet with 16 shots and testing on 10 different datasets without fine-tuning. This experiment examines how well the model transfers to new recognition tasks outside the source domain.

\textbf{Main Results:} Table~\ref{tab:dag} shows the results. After 100 epochs of training, MSGCoOp ($ep100$) reaches an average accuracy of 64.17\% on the target datasets. This is lower than the performance of CLIP (65.12\%), CoCoOp (65.74\%), and KgCoOp (65.51\%). The model appears to overfit to ImageNet after longer training, and the prompt ensemble fails to provide features that transfer well to other datasets. When the target classes match those of ImageNet, as in the cross-domain experiments in Section~\ref{sec: xdo_gen}, the learned prompts work better and generalize to these classes.

\textbf{Early Stopping Performance:} Interestingly, at 5 epochs, MSGCoOp ($ep5$) achieves an average accuracy of 65.92\%, which is higher than all baselines. It surpasses CoCoOp by 0.18\% and KgCoOp by 0.41\%. The multi-prompt semantic-guided learning helps the model use diverse semantic cues early in training. This leads to better generalization before overfitting happens. On datasets like Caltech101 (94.66\%) and SUN397 (67.66\%), MSGCoOp also performs better than the baselines.

\subsection{Ablation Study}
\label{sec:ablation}

We perform ablation studies to assess the contribution of each component in MSGCoOp. Specifically, we examine the effect of the multi-prompt ensemble size ($N$), the semantic guidance from LLM-generated descriptions ($\mathcal{L}_{\mathrm{sg}}$), and the diversity regularization loss ($\mathcal{L}_{\mathrm{div}}$). All experiments are conducted under the 16-shot base-to-novel generalization setting, and results are averaged across the 11 datasets unless noted otherwise.

\subsubsection{Effect of Multi-Prompt Ensemble Size}

\begin{figure}[ht]
    \centering
    \includegraphics[width=0.5\textwidth]{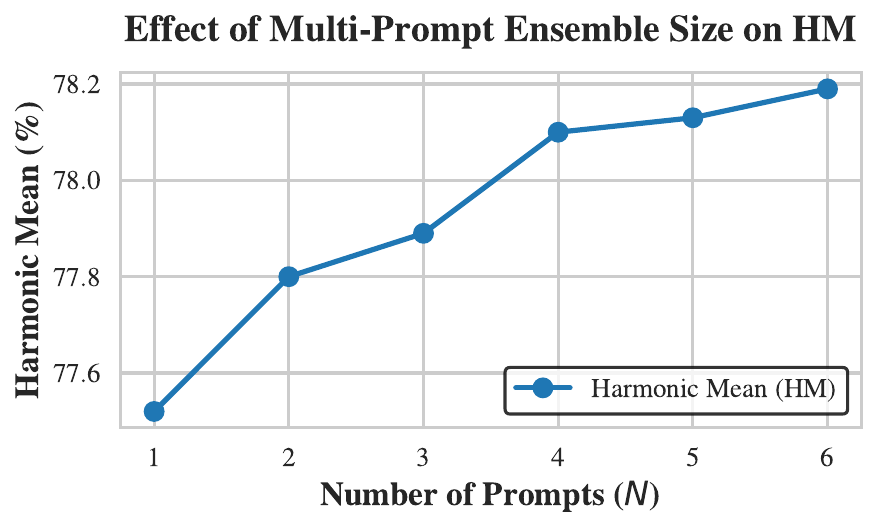}
    \caption{Impact of ensemble size $N$ on harmonic mean (HM).}
    \label{fig:abl_N}
\end{figure}

Our framework employs a multi-prompt ensemble strategy to improve generalization. To evaluate its effectiveness and select a suitable configuration, we vary the number of parallel prompts $N \in \{1, 2, 3, 4, 5, 6\}$ and report results in Figure~\ref{fig:abl_N}. When $N=1$, the method reduces to a single-prompt baseline with semantic guidance, and the diversity regularization term $\mathcal{L}_{\mathrm{div}}$ is disabled.

The results demonstrate consistent gains as $N$ increases. Moving from $N=1$ to $N=6$ yields a +0.67\% improvement in HM, indicating the benefit of leveraging multiple prompts to capture richer semantic cues. Performance peaks at $N=6$ (78.19\% HM), achieving strong generalization on novel classes.

However, gains diminish beyond $N=4$, with $N=6$ only slightly outperforming $N=4$ (+0.09\% HM). Since larger ensembles need higher computational costs with limited additional benefits, we adopt $N=4$ as the default configuration, balancing accuracy and efficiency.

\subsubsection{Effect of LLM-Generated Class Descriptions}

\begin{table}[h]
\centering
\caption{Ablation of LLM-generated semantic guide}
\label{tab:abl_LLM}
\begin{tabular}{l c}
\toprule
Method & Accuracy (\%) \\
\midrule
KgCoOp                   & 77.00 \\
MSGCoOp (hand-crafted)   & 77.28 \\
MSGCoOp (llm-generated)  & 78.10 \\
\bottomrule
\end{tabular}
\end{table}

In all experiments, we set the semantic guidance weight to $\lambda_{\mathrm{sg}} = 8.0$, following the configuration of KgCoOp~\cite{yao2023visual}. 

To evaluate the impact of LLM-generated semantic descriptions, we compare our method with a variant in which the rich descriptions are replaced by simple class templates for semantic guidance. So we replace the semantic reference templates with hand-crafted prompts (e.g., ``a photo of \{cls\}''). 

As reported in Table~\ref{tab:abl_LLM}, our full model attains an accuracy of 78.10\%, surpassing the hand-crafted variant (77.28\%) by a notable margin of +0.82\%. Notably, the hand-crafted variant itself improves over the KgCoOp baseline by +0.28\%, which can be attributed to the benefit of the multi-prompt ensemble.

\subsubsection{Effect of Diversity Regularization}

\begin{figure}[ht]
    \centering
    \includegraphics[width=0.5\textwidth]{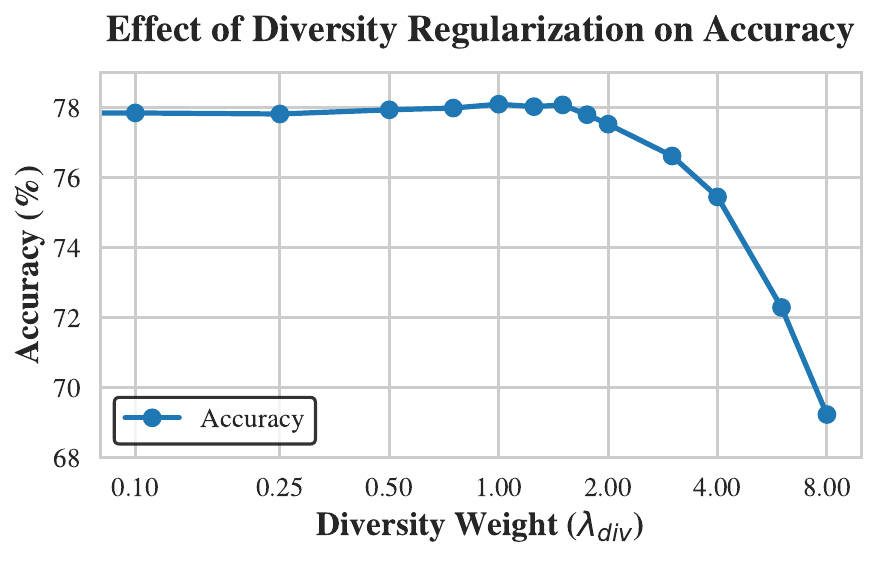}
    \caption{Impact of diversity regularization strength $\lambda_{\mathrm{div}}$ on accuracy.}
    \label{fig:abl_div}
\end{figure}

To evaluate the effect of diversity regularization, we test different regularization weight $\lambda_{\mathrm{div}}$ and get the results in Figure~\ref{fig:abl_div}. This experiment evaluates how encouraging prompt diversity impacts generalization.

As $\lambda_{\mathrm{div}}$ increases from $0.0$ to $1.0$, performance improves steadily, reaching a peak accuracy of 78.10\% at $\lambda_{\mathrm{div}}=1.0$. This highlights the benefit of promoting diverse prompt embeddings, which capture richer semantic variations and enhance robustness to domain shifts.

However, further increasing $\lambda_{\mathrm{div}}$ beyond $1.0$ leads to a gradual decline in accuracy. Excessive regularization destabilizes prompt learning, potentially hindering alignment with target concepts. For example, at $\lambda_{\mathrm{div}}=6.0$ and $\lambda_{\mathrm{div}}=8.0$, accuracy drops to 72.29\% and 69.23\%.

These results suggest that properly tuning the diversity weight is critical for balancing prompt diversity and semantic alignment. We therefore adopt $\lambda_{\mathrm{div}}=1.0$ as the default setting in experiments.

\section{Conclusion}
\label{sec:conclusion}

We present MSGCoOp, a textual prompt learning framework for vision-language models. MSGCoOp improves few-shot generalization by using multiple prompts, semantic class descriptions from large language models, and a diversity regularization strategy. The model uses parallel context vectors and an ensemble fusion approach. This design helps MSGCoOp capture more semantic information without increasing computation much. Experiments show that MSGCoOp achieves higher harmonic mean accuracy and better robustness to domain shifts compared to recent methods. Ablation studies confirm the effectiveness of semantic guidance and prompt diversity. In future work, we will investigate incorporating visual branch features to further improve prompt representations and support adaptation to new tasks.

\newpage
{\small
\bibliographystyle{ieeenat_fullname}
\bibliography{main}

\begin{thebibliography}{38}
\providecommand{\natexlab}[1]{#1}
\providecommand{\url}[1]{\texttt{#1}}
\expandafter\ifx\csname urlstyle\endcsname\relax
  \providecommand{\doi}[1]{doi: #1}\else
  \providecommand{\doi}{doi: \begingroup \urlstyle{rm}\Url}\fi

\bibitem[Achiam et~al.(2023)Achiam, Adler, Agarwal, Ahmad, Akkaya, Aleman, Almeida, Altenschmidt, Altman, Anadkat, et~al.]{achiam2023gpt}
Josh Achiam, Steven Adler, Sandhini Agarwal, Lama Ahmad, Ilge Akkaya, Florencia~Leoni Aleman, Diogo Almeida, Janko Altenschmidt, Sam Altman, Shyamal Anadkat, et~al.
\newblock Gpt-4 technical report.
\newblock \emph{arXiv preprint arXiv:2303.08774}, 2023.

\bibitem[Alayrac et~al.(2022)Alayrac, Donahue, Luc, Miech, Barr, Hasson, Lenc, Mensch, Millican, Reynolds, et~al.]{alayrac2022flamingo}
Jean-Baptiste Alayrac, Jeff Donahue, Pauline Luc, Antoine Miech, Iain Barr, Yana Hasson, Karel Lenc, Arthur Mensch, Katherine Millican, Malcolm Reynolds, et~al.
\newblock Flamingo: a visual language model for few-shot learning.
\newblock \emph{Advances in neural information processing systems}, 35:\penalty0 23716--23736, 2022.

\bibitem[Bossard et~al.(2014)Bossard, Guillaumin, and Gool]{bossard2014food}
Lukas Bossard, Matthieu Guillaumin, and Luc~Van Gool.
\newblock Food-101--mining discriminative components with random forests.
\newblock In \emph{ECCV}, pages 446--461. Springer, 2014.

\bibitem[Brown et~al.(2020)Brown, Mann, Ryder, Subbiah, Kaplan, Dhariwal, Neelakantan, Shyam, Sastry, Askell, et~al.]{brown2020language}
Tom Brown, Benjamin Mann, Nick Ryder, Melanie Subbiah, Jared~D Kaplan, Prafulla Dhariwal, Arvind Neelakantan, Pranav Shyam, Girish Sastry, Amanda Askell, et~al.
\newblock Language models are few-shot learners.
\newblock \emph{Advances in neural information processing systems}, 33:\penalty0 1877--1901, 2020.

\bibitem[Cimpoi et~al.(2014)Cimpoi, Maji, Kokkinos, Mohamed, and Vedaldi]{cimpoi2014describing}
Mircea Cimpoi, Subhransu Maji, Iasonas Kokkinos, Sammy Mohamed, and Andrea Vedaldi.
\newblock Describing textures in the wild.
\newblock In \emph{CVPR}, pages 3606--3613, 2014.

\bibitem[Deng et~al.(2009)Deng, Dong, Socher, Li, Li, and Fei-Fei]{deng2009imagenet}
Jia Deng, Wei Dong, Richard Socher, Li-Jia Li, Kai Li, and Li Fei-Fei.
\newblock Imagenet: A large-scale hierarchical image database.
\newblock In \emph{CVPR}, pages 248--255. Ieee, 2009.

\bibitem[Fei-Fei et~al.(2004)Fei-Fei, Fergus, and Perona]{fei2004learning}
Li Fei-Fei, Rob Fergus, and Pietro Perona.
\newblock Learning generative visual models from few training examples: An incremental bayesian approach tested on 101 object categories.
\newblock In \emph{2004 conference on computer vision and pattern recognition workshop}, pages 178--178. IEEE, 2004.

\bibitem[Gao et~al.(2024)Gao, Geng, Zhang, Ma, Fang, Zhang, Li, and Qiao]{gao2024clip}
Peng Gao, Shijie Geng, Renrui Zhang, Teli Ma, Rongyao Fang, Yongfeng Zhang, Hongsheng Li, and Yu Qiao.
\newblock Clip-adapter: Better vision-language models with feature adapters.
\newblock \emph{International Journal of Computer Vision}, 132\penalty0 (2):\penalty0 581--595, 2024.

\bibitem[Helber et~al.(2019)Helber, Bischke, Dengel, and Borth]{helber2019eurosat}
Patrick Helber, Benjamin Bischke, Andreas Dengel, and Damian Borth.
\newblock Eurosat: A novel dataset and deep learning benchmark for land use and land cover classification.
\newblock \emph{IEEE Journal of Selected Topics in Applied Earth Observations and Remote Sensing}, 12\penalty0 (7):\penalty0 2217--2226, 2019.

\bibitem[Hendrycks et~al.(2021{\natexlab{a}})Hendrycks, Basart, Mu, Kadavath, Wang, Dorundo, Desai, Zhu, Parajuli, Guo, et~al.]{hendrycks2021many}
Dan Hendrycks, Steven Basart, Norman Mu, Saurav Kadavath, Frank Wang, Evan Dorundo, Rahul Desai, Tyler Zhu, Samyak Parajuli, Mike Guo, et~al.
\newblock The many faces of robustness: A critical analysis of out-of-distribution generalization.
\newblock In \emph{ICCV}, pages 8340--8349, 2021{\natexlab{a}}.

\bibitem[Hendrycks et~al.(2021{\natexlab{b}})Hendrycks, Zhao, Basart, Steinhardt, and Song]{hendrycks2021natural}
Dan Hendrycks, Kevin Zhao, Steven Basart, Jacob Steinhardt, and Dawn Song.
\newblock Natural adversarial examples.
\newblock In \emph{CVPR}, pages 15262--15271, 2021{\natexlab{b}}.

\bibitem[Jia et~al.(2021)Jia, Yang, Xia, Chen, Parekh, Pham, Le, Sung, Li, and Duerig]{jia2021scaling}
Chao Jia, Yinfei Yang, Ye Xia, Yi-Ting Chen, Zarana Parekh, Hieu Pham, Quoc Le, Yun-Hsuan Sung, Zhen Li, and Tom Duerig.
\newblock Scaling up visual and vision-language representation learning with noisy text supervision.
\newblock In \emph{International Conference on Machine Learning}, pages 4904--4916. PMLR, 2021.

\bibitem[Khattak et~al.(2023)Khattak, Rasheed, Maaz, Khan, and Khan]{khattak2023maple}
Muhammad~Uzair Khattak, Hanoona Rasheed, Muhammad Maaz, Salman Khan, and Fahad~Shahbaz Khan.
\newblock Maple: Multi-modal prompt learning.
\newblock In \emph{Proceedings of the IEEE/CVF conference on computer vision and pattern recognition}, pages 19113--19122, 2023.

\bibitem[Kim et~al.(2024)Kim, Kim, and Lee]{kim2024aapl}
Gahyeon Kim, Sohee Kim, and Seokju Lee.
\newblock Aapl: Adding attributes to prompt learning for vision-language models.
\newblock In \emph{Proceedings of the IEEE/CVF conference on computer vision and pattern recognition}, pages 1572--1582, 2024.

\bibitem[Krause et~al.(2013)Krause, Stark, Deng, and Fei-Fei]{krause20133d}
Jonathan Krause, Michael Stark, Jia Deng, and Li Fei-Fei.
\newblock 3d object representations for fine-grained categorization.
\newblock In \emph{ICCV}, pages 554--561, 2013.

\bibitem[Li et~al.(2023)Li, Li, Savarese, and Hoi]{li2023blip}
Junnan Li, Dongxu Li, Silvio Savarese, and Steven Hoi.
\newblock Blip-2: Bootstrapping language-image pre-training with frozen image encoders and large language models.
\newblock In \emph{International conference on machine learning}, pages 19730--19742. PMLR, 2023.

\bibitem[Liu et~al.(2024)Liu, Feng, Xue, Wang, Wu, Lu, Zhao, Deng, Zhang, Ruan, et~al.]{liu2024deepseek}
Aixin Liu, Bei Feng, Bing Xue, Bingxuan Wang, Bochao Wu, Chengda Lu, Chenggang Zhao, Chengqi Deng, Chenyu Zhang, Chong Ruan, et~al.
\newblock Deepseek-v3 technical report.
\newblock \emph{arXiv preprint arXiv:2412.19437}, 2024.

\bibitem[Liu et~al.(2023{\natexlab{a}})Liu, Li, Wu, and Lee]{liu2023llava}
Haotian Liu, Chunyuan Li, Qingyang Wu, and Yong~Jae Lee.
\newblock Visual instruction tuning, 2023{\natexlab{a}}.

\bibitem[Liu et~al.(2023{\natexlab{b}})Liu, Son, Yang, Liu, Gao, Lee, and Li]{liu2023learning}
Haotian Liu, Kilho Son, Jianwei Yang, Ce Liu, Jianfeng Gao, Yong~Jae Lee, and Chunyuan Li.
\newblock Learning customized visual models with retrieval-augmented knowledge.
\newblock In \emph{Proceedings of the IEEE/CVF Conference on Computer Vision and Pattern Recognition}, pages 15148--15158, 2023{\natexlab{b}}.

\bibitem[Maji et~al.(2013)Maji, Rahtu, Kannala, Blaschko, and Vedaldi]{maji2013fine}
Subhransu Maji, Esa Rahtu, Juho Kannala, Matthew Blaschko, and Andrea Vedaldi.
\newblock Fine-grained visual classification of aircraft.
\newblock \emph{arXiv preprint arXiv:1306.5151}, 2013.

\bibitem[Mu et~al.(2022)Mu, Kirillov, Wagner, and Xie]{mu2022slip}
Norman Mu, Alexander Kirillov, David Wagner, and Saining Xie.
\newblock Slip: Self-supervision meets language-image pre-training.
\newblock In \emph{European conference on computer vision}, pages 529--544. Springer, 2022.

\bibitem[Nilsback and Zisserman(2008)]{nilsback2008automated}
Maria-Elena Nilsback and Andrew Zisserman.
\newblock Automated flower classification over a large number of classes.
\newblock In \emph{2008 Sixth Indian Conference on Computer Vision, Graphics \& Image Processing}, pages 722--729. IEEE, 2008.

\bibitem[Parkhi et~al.(2012)Parkhi, Vedaldi, Zisserman, and Jawahar]{parkhi2012cats}
Omkar~M Parkhi, Andrea Vedaldi, Andrew Zisserman, and CV Jawahar.
\newblock Cats and dogs.
\newblock In \emph{2012 IEEE conference on computer vision and pattern recognition}, pages 3498--3505. IEEE, 2012.

\bibitem[Radford et~al.(2021)Radford, Kim, Hallacy, Ramesh, Goh, Agarwal, Sastry, Askell, Mishkin, Clark, et~al.]{radford2021learning}
Alec Radford, Jong~Wook Kim, Chris Hallacy, Aditya Ramesh, Gabriel Goh, Sandhini Agarwal, Girish Sastry, Amanda Askell, Pamela Mishkin, Jack Clark, et~al.
\newblock Learning transferable visual models from natural language supervision.
\newblock In \emph{International Conference on Machine Learning}, pages 8748--8763. PMLR, 2021.

\bibitem[Recht et~al.(2019)Recht, Roelofs, Schmidt, and Shankar]{recht2019imagenet}
Benjamin Recht, Rebecca Roelofs, Ludwig Schmidt, and Vaishaal Shankar.
\newblock Do imagenet classifiers generalize to imagenet?
\newblock In \emph{International Conference on Machine Learning}, pages 5389--5400. PMLR, 2019.

\bibitem[Schick and Sch{\"u}tze(2020)]{schick2020exploiting}
Timo Schick and Hinrich Sch{\"u}tze.
\newblock Exploiting cloze questions for few shot text classification and natural language inference.
\newblock \emph{arXiv preprint arXiv:2001.07676}, 2020.

\bibitem[Shen et~al.(2022)Shen, Li, Hu, Xie, Yang, Zhang, Gan, Wang, Yuan, Liu, et~al.]{shen2022k}
Sheng Shen, Chunyuan Li, Xiaowei Hu, Yujia Xie, Jianwei Yang, Pengchuan Zhang, Zhe Gan, Lijuan Wang, Lu Yuan, Ce Liu, et~al.
\newblock K-lite: Learning transferable visual models with external knowledge.
\newblock \emph{Advances in Neural Information Processing Systems}, 35:\penalty0 15558--15573, 2022.

\bibitem[Song et~al.(2023)Song, Xue, Wang, Sun, Ge, Shan, et~al.]{song2023meta}
Lin Song, Ruoyi Xue, Hang Wang, Hongbin Sun, Yixiao Ge, Ying Shan, et~al.
\newblock Meta-adapter: An online few-shot learner for vision-language model.
\newblock \emph{Advances in Neural Information Processing Systems}, 36:\penalty0 55361--55374, 2023.

\bibitem[Soomro et~al.(2012)Soomro, Zamir, and Shah]{soomro2012ucf101}
Khurram Soomro, Amir~Roshan Zamir, and Mubarak Shah.
\newblock Ucf101: A dataset of 101 human actions classes from videos in the wild.
\newblock \emph{arXiv preprint arXiv:1212.0402}, 2012.

\bibitem[Wang et~al.(2019)Wang, Ge, Lipton, and Xing]{wang2019learning}
Haohan Wang, Songwei Ge, Zachary Lipton, and Eric~P Xing.
\newblock Learning robust global representations by penalizing local predictive power.
\newblock In \emph{NeurIPS}, 2019.

\bibitem[Xiao et~al.(2010)Xiao, Hays, Ehinger, Oliva, and Torralba]{xiao2010sun}
Jianxiong Xiao, James Hays, Krista~A Ehinger, Aude Oliva, and Antonio Torralba.
\newblock Sun database: Large-scale scene recognition from abbey to zoo.
\newblock In \emph{2010 IEEE computer society conference on computer vision and pattern recognition}, pages 3485--3492. IEEE, 2010.

\bibitem[Yang et~al.(2022)Yang, Li, Zhang, Xiao, Liu, Yuan, and Gao]{yang2022unified}
Jianwei Yang, Chunyuan Li, Pengchuan Zhang, Bin Xiao, Ce Liu, Lu Yuan, and Jianfeng Gao.
\newblock Unified contrastive learning in image-text-label space.
\newblock In \emph{Proceedings of the IEEE/CVF conference on computer vision and pattern recognition}, pages 19163--19173, 2022.

\bibitem[Yao et~al.(2023)Yao, Zhang, and Xu]{yao2023visual}
Hantao Yao, Rui Zhang, and Changsheng Xu.
\newblock Visual-language prompt tuning with knowledge-guided context optimization.
\newblock In \emph{Proceedings of the IEEE/CVF conference on computer vision and pattern recognition}, pages 6757--6767, 2023.

\bibitem[Yao et~al.(2021)Yao, Huang, Hou, Lu, Niu, Xu, Liang, Li, Jiang, and Xu]{yao2021filip}
Lewei Yao, Runhui Huang, Lu Hou, Guansong Lu, Minzhe Niu, Hang Xu, Xiaodan Liang, Zhenguo Li, Xin Jiang, and Chunjing Xu.
\newblock Filip: Fine-grained interactive language-image pre-training.
\newblock \emph{arXiv preprint arXiv:2111.07783}, 2021.

\bibitem[Yuan et~al.(2021)Yuan, Chen, Chen, Codella, Dai, Gao, Hu, Huang, Li, Li, et~al.]{yuan2021florence}
Lu Yuan, Dongdong Chen, Yi-Ling Chen, Noel Codella, Xiyang Dai, Jianfeng Gao, Houdong Hu, Xuedong Huang, Boxin Li, Chunyuan Li, et~al.
\newblock Florence: A new foundation model for computer vision.
\newblock \emph{arXiv preprint arXiv:2111.11432}, 2021.

\bibitem[Zhou et~al.(2022{\natexlab{a}})Zhou, Yang, Loy, and Liu]{zhou2022conditional}
Kaiyang Zhou, Jingkang Yang, Chen~Change Loy, and Ziwei Liu.
\newblock Conditional prompt learning for vision-language models.
\newblock In \emph{Proceedings of the IEEE/CVF conference on computer vision and pattern recognition}, pages 16816--16825, 2022{\natexlab{a}}.

\bibitem[Zhou et~al.(2022{\natexlab{b}})Zhou, Yang, Loy, and Liu]{zhou2022learning}
Kaiyang Zhou, Jingkang Yang, Chen~Change Loy, and Ziwei Liu.
\newblock Learning to prompt for vision-language models.
\newblock \emph{International Journal of Computer Vision}, 130\penalty0 (9):\penalty0 2337--2348, 2022{\natexlab{b}}.

\bibitem[Zhu et~al.(2023)Zhu, Niu, Han, Wu, and Zhang]{zhu2023prompt}
Beier Zhu, Yulei Niu, Yucheng Han, Yue Wu, and Hanwang Zhang.
\newblock Prompt-aligned gradient for prompt tuning.
\newblock In \emph{Proceedings of the IEEE/CVF international conference on computer vision}, pages 15659--15669, 2023.

\end{thebibliography}
}


\end{document}